\documentclass{article} 
\usepackage{iclr2017_conference,times}
\usepackage{hyperref}
\usepackage{url}
\usepackage{array}
\usepackage{tabularx, booktabs}
\usepackage{soul}
\usepackage{subfig}
\usepackage{color}
\usepackage{graphicx}
\usepackage{mathtools}
\usepackage{authblk}

\graphicspath{ {images/} }

\newcommand{\eat}[1]{\ignorespaces}

\pdfoutput=1

\usepackage{amsmath}
\usepackage{tikz}
\usepackage[font={footnotesize}]{caption}
\tikzset{font=\scriptsize}
\usetikzlibrary{patterns}

\newcommand{\ignore}[1]{}
\newcommand{\argmax}{\operatornamewithlimits{argmax}}

\definecolor{g-red}{HTML}{DB4437}
\definecolor{g-blue}{HTML}{4285F4}
\definecolor{g-green}{HTML}{0F9D58}
\definecolor{g-yellow}{HTML}{F4B400}
\definecolor{g-orange}{HTML}{FF9800}
\definecolor{g-grey}{HTML}{9E9E9E}

\newcommand{\ques}{\mathbf{q}}
\newcommand{\pass}{\mathbf{p}}
\newcommand{\ans}{\mathbf{a}}
\newcommand{\cand}{\mathbf{A}}
\newcommand{\start}{\mathit{start}}
\newcommand{\finish}{\mathit{end}}
\newcommand{\bilstm}{BiLSTM}

\newcommand{\squad}{{\sc SQuAD}}
\newcommand{\ourmodel}{\textsc{RaSoR}}

\newcommand\layerbox[4]{
\draw[rounded corners] (#2, #3) rectangle (#2 + #1 * #4, #3 + #1 * 1);
}

\newcommand\layercolorbox[5]{
\draw[rounded corners, fill=#5] (#2, #3) rectangle (#2 + #1 * #4, #3 + #1 * 1);
}
\newcommand\layercomponent[5]{
\filldraw[fill=#5] (#2 + #1 * #4 - #1 * 0.5, #3 + #1 * 0.5) circle (#1 * 0.4);
}
\newcommand\layer[5][0.4] {
\layerbox{#1}{#2}{#3}{#4}
\foreach \x in {1, ..., #4}{
  \layercomponent{#1}{#2}{#3}{\x}{#5}
}
}

\title{Learning Recurrent Span Representations for Extractive Question Answering}

\author[$\dagger$]{Kenton Lee}
\author[$\star$]{Shimi Salant}
\author[$\ddag$]{Tom Kwiatkowksi}
\author[$\ddag$]{Ankur Parikh}
\author[$\ddag$]{Dipanjan Das}
\author[$\star$]{Jonathan Berant}
\affil[ ]{\texttt{kentonl@cs.washington.edu, shimonsalant@mail.tau.ac.il}}
\affil[ ]{\texttt{\{tomkwiat, aparikh, dipanjand\}@google.com, joberant@cs.tau.ac.il}}
\affil[ ]{ }
\affil[$\dagger$]{University of Washington, Seattle, USA}
\affil[$\star$]{Tel-Aviv University, Tel-Aviv, Israel}
\affil[$\ddag$]{Google Research, New York, USA}

\begin{document}
\maketitle

\begin{abstract}

The reading comprehension task, that asks questions about a given evidence document, is a central problem in natural language understanding.
Recent formulations of this task have typically focused on \emph{answer selection} from a set of candidates pre-defined manually or through the use of an external NLP pipeline.
However, \cite{rajpurkar:2016} recently released the \squad~dataset in which the answers can be arbitrary strings from the supplied text.
In this paper, we focus on this \emph{answer extraction} task, presenting a novel model architecture that efficiently builds fixed length representations of all spans in the evidence document with a recurrent network.
We show that scoring explicit span representations significantly improves performance over other approaches that factor the prediction into separate predictions about words or start and end markers.
Our approach improves upon the best published results of \cite{wang2016machine} by $5\%$  and decreases the error of \citeauthor{rajpurkar:2016}'s baseline by $>50\%$.
\end{abstract}


\section{Introduction}
A primary goal of natural language processing is to develop systems that can answer questions about the contents of documents.
The reading comprehension task is of practical interest -- we want computers to be able to read the world's text and then answer our questions -- and, since we believe it requires deep language understanding, it has also become a flagship task in NLP research.

A number of reading comprehension datasets have been developed that focus on answer selection from a small set of alternatives defined by annotators~\citep{richardson2013mctest} or existing NLP pipelines that cannot be trained end-to-end~\citep{hill2015goldilocks, hermann2015teaching}.
Subsequently, the models proposed for this task have tended to make use of the limited set of candidates, basing their predictions on mention-level attention weights~\citep{hermann2015teaching}, or centering classifiers~\citep{chen2016thorough}, or network memories~\citep{hill2015goldilocks} on candidate locations.

Recently, \cite{rajpurkar:2016} released the less restricted \squad~dataset\footnote{\url{http://stanford-qa.com}} that does not place any constraints on the set of allowed answers, other than that they should be drawn from the evidence document.
\citeauthor{rajpurkar:2016} proposed a baseline system that chooses answers from the constituents identified by an existing syntactic parser.
This allows them to prune the $O(N^2)$ answer candidates in each document of length $N$, but it also effectively renders $20.7\%$ of all questions unanswerable.

Subsequent work by \cite{wang2016machine} significantly improve upon this baseline by using an end-to-end neural network architecture to identify answer spans by labeling either individual words, or the start and end of the answer span.
Both of these methods do not make independence assumptions about substructures, but they are susceptible to search errors due to greedy training and decoding.

In contrast, here we argue that it is beneficial to simplify the decoding procedure by enumerating all possible answer spans. 
By explicitly representing each answer span, our model can be globally normalized during training and decoded exactly during evaluation.
A naive approach to building the $O(N^2)$ spans of up to length $N$ would require a network that is cubic in size with respect to the passage length, and such a network would be untrainable.
To overcome this, we present a novel neural architecture called \ourmodel~that builds fixed-length span representations, \textit{reusing} recurrent computations for shared substructures.  We demonstrate that directly classifying each of the competing spans, and training with global normalization over all possible spans, leads to a significant increase in performance.
In our experiments, we show an increase in performance over \cite{wang2016machine} of $5\%$ in terms of exact match to a reference answer, and  $3.6\%$ in terms of predicted answer F1 with respect to the reference. On both of these metrics, we close the gap between \citeauthor{rajpurkar:2016}'s baseline and the human-performance upper-bound by $>50\%$.
\section{Extractive Question Answering}
\subsection{Task Definition}
Extractive question answering systems take as input a question $\ques=\{q_0, \dots, q_n\}$ and a passage of text $\pass=\{p_0,\dots,p_m\}$ from which they predict a single answer span $\ans=\langle a_{\start}, a_{\finish}\rangle$, represented as a pair of indices into $\pass$.
Machine learned extractive question answering systems, such as the one presented here, learn a predictor function $f(\ques, \pass) \rightarrow \ans$ from a training dataset of $\langle \ques, \pass, \ans \rangle$ triples.


\subsection{Related Work}
For the \squad~dataset, the original paper from \cite{rajpurkar:2016} implemented a linear model with sparse features based on $n$-grams and part-of-speech tags present in the question and the candidate answer.
Other than lexical features, they also used syntactic information in the form of dependency paths to extract more general features.
They set a strong baseline for following work and also presented an in depth analysis, showing that lexical and syntactic features contribute most strongly to their model's performance.  
Subsequent work by \cite{wang2016machine} use an end-to-end neural network method that uses a Match-LSTM to model the question and the passage, and uses pointer networks \citep{vinyals:2015} to extract the answer span from the passage.  
This model resorts to greedy decoding and falls short in terms of performance compared to our model (see Section~\ref{sec:results} for more detail). 
While we only compare to published baselines, there are other unpublished competitive systems on the \squad~leaderboard, as listed in footnote~\ref{ftnt:leaderboard}.

A task that is closely related to extractive question answering is the Cloze task \citep{taylor:1953}, in which the goal is to predict a concealed span from a declarative sentence given a passage of supporting text.
Recently, \cite{hermann2015teaching} presented a Cloze dataset in which the task is to predict the correct entity in an incomplete sentence given an abstractive summary of a news article.  \citeauthor{hermann2015teaching} also present various neural architectures to solve the problem.
Although this dataset is large and varied in domain, recent analysis by \cite{chen2016thorough} shows that simple models can achieve close to the human upper bound.
As noted by the authors of the \squad~paper, the annotated answers in the \squad~dataset are often spans that include non-entities and can be longer phrases, unlike the Cloze datasets, thus making the task more challenging.

Another, more traditional line of work has focused on extractive question answering on sentences, where the task is to extract a sentence from a document, given a question.
Relevant datasets include datasets from the annual TREC evaluations \citep{voorhees:2000} and WikiQA \citep{yang:2015}, where the latter dataset specifically focused on Wikipedia passages.
There has been a line of interesting recent publications using neural architectures, focused on this variety of extractive question answering \citep[\textit{inter alia}]{tymoshenko:2016,wang:2016}.
These methods model the question and a candidate answer sentence, but do not focus on possible candidate answer \textit{spans} that may contain the answer to the given question.
In this work, we focus on the more challenging problem of extracting the precise answer span.


\section{Model}
\label{sec:model}
We propose a model architecture called \ourmodel\footnote{An abbreviation for \underline{R}ecurrent \underline{S}pan \underline{R}epresentations, pronounced as \textit{razor}.} illustrated in Figure~{\ref{fig:span}}, that explicitly computes embedding representations for candidate answer spans.
In most structured prediction problems (e.g. sequence labeling or parsing), the number of possible output structures is exponential in the input length, and computing representations for every candidate is prohibitively expensive.
However, we exploit the simplicity of our task, where we can trivially and tractably enumerate all candidates.
This facilitates an expressive model that computes joint representations of every answer span, that can be globally normalized during learning.

In order to compute these span representations, we must aggregate information from the passage and the question for every answer candidate. For the example in Figure~\ref{fig:span}, \ourmodel~computes an embedding for the candidate answer spans: \textit{fixed to}, \textit{fixed to the}, \textit{to the}, etc. A naive approach for these aggregations would require a network that is cubic in size with respect to the passage length.
Instead, our model reduces this to a quadratic size by reusing recurrent computations for shared substructures (i.e. common passage words) from different spans.

Since the choice of answer span depends on the original question, we must incorporate this information into the computation of the span representation. We model this by augmenting the passage word embeddings with additional embedding representations of the question.

In this section, we motivate and describe the architecture for \ourmodel~in a top-down manner.

\subsection{Scoring Answer Spans}
The goal of our extractive question answering system is to predict the single best answer span among all candidates from the passage $\pass$, denoted as $\cand(\pass)$.
Therefore, we define a probability distribution over all possible answer spans given the question $\ques$ and passage $\pass$, and the predictor function finds the answer span with the maximum likelihood:
\begin{align}
f(\ques, \pass) &\vcentcolon= \argmax_{\ans \in \cand(\pass)} P(\ans \mid \ques, \pass)
\end{align}
One might be tempted to introduce independence assumptions that would enable cheaper decoding.
For example, this distribution can be modeled as (1) a product of conditionally independent distributions (binary) for every word or (2) a product of conditionally independent distributions (over words) for the start and end indices of the answer span.
However, we show in Section~\ref{sec:ablations} that such independence assumptions hurt the accuracy of the model, and instead we only assume a fixed-length representation $h_{\ans}$ of each candidate span that is scored and normalized with a softmax layer (\textbf{Span score} and \textbf{Softmax} in Figure~\ref{fig:span}):
\begin{align}
s_{\ans} &= w_a \cdot \textsc{ffnn}(h_{\ans}) & \ans \in \cand(\pass)\\
P(\ans \mid \ques, \pass) &= \frac{\exp(s_{\ans})}{\sum_{\ans' \in \cand(\pass)} \exp(s_{\ans'})} & \ans \in \cand(\pass)
\end{align}
where $\textsc{ffnn}(\cdot)$ denotes a fully connected feed-forward neural network that provides a non-linear mapping of its input embedding.

\subsection{\ourmodel: Recurrent Span Representation}
The previously defined probability distribution depends on the answer span representations, $h_{\ans}$.
When computing $h_{\ans}$, we assume access to representations of individual passage words that have been augmented with a representation of the question.
We denote these question-focused passage word embeddings as $\{p^*_1, \dots, p^*_m\}$ and describe their creation in Section~\ref{sec:pstar}.
In order to reuse computation for shared substructures, we use a bidirectional LSTM \citep{lstm} to encode the left and right context of every $p^*_i$ (\textbf{Passage-level \bilstm} in Figure~\ref{fig:span}).
This allows us to simply concatenate the bidirectional LSTM (\bilstm) outputs at the endpoints of a span to jointly encode its inside and outside information (\textbf{Span embedding} in Figure~\ref{fig:span}):
\begin{align}
\{p^{*\prime}_1, \dots, p^{*\prime}_m\} &= \textsc{bilstm}(\{p^*_1, \dots, p^*_m\})\\
h_{\ans} &= [p^{*\prime}_{a_{\start}}, p^{*'}_{a_{\finish}}] & \langle a_{\start}, a_{\finish}\rangle \in \cand(\pass)
\end{align}
where $\textsc{bilstm}(\cdot)$ denotes a \bilstm~over its input embedding sequence and $p^{*\prime}_i$ is the concatenation of forward and backward outputs at time-step $i$. While the visualization in Figure~\ref{fig:span} shows a single layer \bilstm~for simplicity, we use a multi-layer \bilstm~in our experiments. The concatenated output of each layer is used as input for the subsequent layer, allowing the upper layers to depend on the entire passage.

\subsection{Question-focused Passage Word Embedding}
\label{sec:pstar}
Computing the question-focused passage word embeddings $\{p^*_1, \dots, p^*_m\}$ requires integrating question information into the passage.
The architecture for this integration is flexible and likely depends on the nature of the dataset.
For the \squad~dataset, we find that both passage-aligned and passage-independent question representations are effective at incorporating this contextual information, and experiments will show that their benefits are complementary.
To incorporate these question representations, we simply concatenate them with the passage word embeddings (\textbf{Question-focused passage word embedding} in Figure~\ref{fig:span}).

We use fixed pretrained embeddings to represent question and passage words. Therefore, in the following discussion, notation for the words are interchangeable with their embedding representations.

\paragraph{Question-independent passage word embedding} The first component simply looks up the pretrained word embedding for the passage word, $p_i$.

\paragraph{Passage-aligned question representation} In this dataset, the question-passage pairs often contain large lexical overlap or similarity near the correct answer span.
To encourage the model to exploit these similarities, we include a fixed-length representation of the question based on soft-alignments with the passage word. 
The alignments are computed via neural attention~\citep{bahdanau:2014}, and we use the variant proposed by \cite{parikh:2016}, where attention scores are dot products between non-linear mappings of word embeddings.
\begin{align}
  s_{ij} &= \textsc{ffnn}(p_{i}) \cdot \textsc{ffnn}(q_j) & 1 \le j \le n\\
  a_{ij} &= \frac{\exp(s_{ij})}{\sum_{k=1}^{n}\exp(s_{ik})} & 1 \le j \le n\\
  q^{align}_{i} &= \sum_{j=1}^{n} a_{ij} q_j
\end{align}

\paragraph{Passage-independent question representation} We also include a representation of the question that does not depend on the passage and is shared for all passage words.

Similar to the previous question representation, an attention score is computed via a dot-product, except the question word is compared to a universal learned embedding rather any particular passage word.
Additionally, we incorporate contextual information with a \bilstm~before aggregating the outputs using this attention mechanism.

The goal is to generate a coarse-grained summary of the question that depends on word order.
Formally, the passage-independent question representation $q^{indep}$ is computed as follows:
\begin{align}
  \{q'_1, \dots, q'_n\} &= \textsc{bilstm}(\ques)\\
  s_j &= w_q \cdot \textsc{ffnn}(q'_j) & 1 \le j \le n\\
  a_j &= \frac{\exp(s_j)}{\sum_{k=1}^{n}\exp(s_k)} & 1 \le j \le n\\
  q^{indep} &= \sum_{j=1}^{n} a_j q'_j \label{eqn:qsumm}
\end{align}
This representation is a bidirectional generalization of the question representation recently proposed by~\cite{li2016dataset} for a different question-answering task.

Given the above three components, the complete question-focused passage word embedding for $p_i$ is their concatenation: $p^*_i = [p_i, q^{align}_{i}, q^{indep}]$.

\subsection{Learning}
Given the above model specification, learning is straightforward. We simply maximize the log-likelihood of the correct answer candidates and backpropagate the errors end-to-end.
\newcommand\qplayer[3][0.4] {
\layerbox{#1}{#2}{#3}{6}
\layercomponent{#1}{#2}{#3}{1}{g-red}
\layercomponent{#1}{#2}{#3}{2}{g-red}
\layercomponent{#1}{#2}{#3}{3}{g-yellow}
\layercomponent{#1}{#2}{#3}{4}{g-yellow}
\layercomponent{#1}{#2}{#3}{5}{g-blue}
\layercomponent{#1}{#2}{#3}{6}{g-blue}
}
\newcommand\lstm[3] {
\filldraw[fill=#3] (#1+0.3,#2+0.7-0.5) rectangle (#1+0.7,#2+0.7); 
\filldraw[fill=#3] (#1+0.3,#2+0.0-0.5) rectangle (#1+0.7,#2+0.0); 

\draw[-latex, line width=1pt, out=75, in=-75] (#1+0.5,#2+0.7) -- (#1+0.7, #2+1); 
\draw[-latex, line width=1pt, out=150, in=-150, looseness=1.8] (#1+0.5, #2) to (#1+0.3, #2+1); 
\draw[-latex, line width=1pt] (#1+0.5,#2-0.8) -- (#1+0.5, #2-0.5); 
\draw[-latex, line width=1pt, out=30, in=-30, looseness=1.8] (#1+0.5, #2-0.8) to (#1+0.5, #2+0.2); 
}
\newcommand\lstmconnect[3] {
\draw[-latex, line width=1pt] (#1+0.7,#2+0.25+0.7) -- (#1+0.3+#3,#2+0.25+0.7);
\draw[-latex, line width=1pt] (#1+0.3+#3,#2+0.25) -- (#1+0.7,#2+0.25);
}
\newcommand\pword[3] {
\layer{#1+0.1}{#2+1.5}{2}{g-green} 
\lstm{#1}{#2+0.5}{g-green}
\qplayer{#1-0.7}{#2-0.7} 
\node[anchor=east, align=right] at (#1+0.3, #2-0.1) {#3}; 
}
\newcommand\qword[3] {
\draw[-latex, line width=1pt] (#1+0.5, #2+1.9) -- (1.5, -1.2-0.2); 
\layer{#1+0.1}{#2+1.5}{2}{g-blue} 
\lstm{#1}{#2+0.5}{g-blue}
\layer{#1+0.1}{#2-0.7}{2}{g-yellow} 
\node[anchor=east, align=right] at (#1+0.5, #2-0.15) {#3}; 
\draw[-latex, line width=1pt] (#1+0.5, #2-0.7) -- (2, -5+0.2); 
}
\newcommand\pspan[8] {
\draw[-latex, line width=1pt, in=#7, out=#8] (#1+0.55,5.3+#2) to (3.15+1+#6, 4.9); 
\layer{#1+0.35}{4.9+#2}{1}{g-grey} 
\draw[-latex, line width=1pt] (#1+0.55,4.5+#2) -- (#1+0.55, 4.9+#2); 
\layer{#1+0.15}{4.1+#2}{2}{white} 
\draw[-latex, line width=1pt] (#1+0.55,3.4+#2) -- (#1+0.55, 4.1+#2); 
\node[anchor=east, align=right] at (#1+0.55, 3.8+#2) {#5}; 
\layer{#1-0.25}{3+#2}{4}{g-orange} 
\draw[-latex, line width=1pt] (#3+0.5,1.9) -- (#1+0.2, 3+#2); 
\draw[-latex, line width=1pt] (#4+0.5,1.9) -- (#1+0.9, 3+#2); 
}

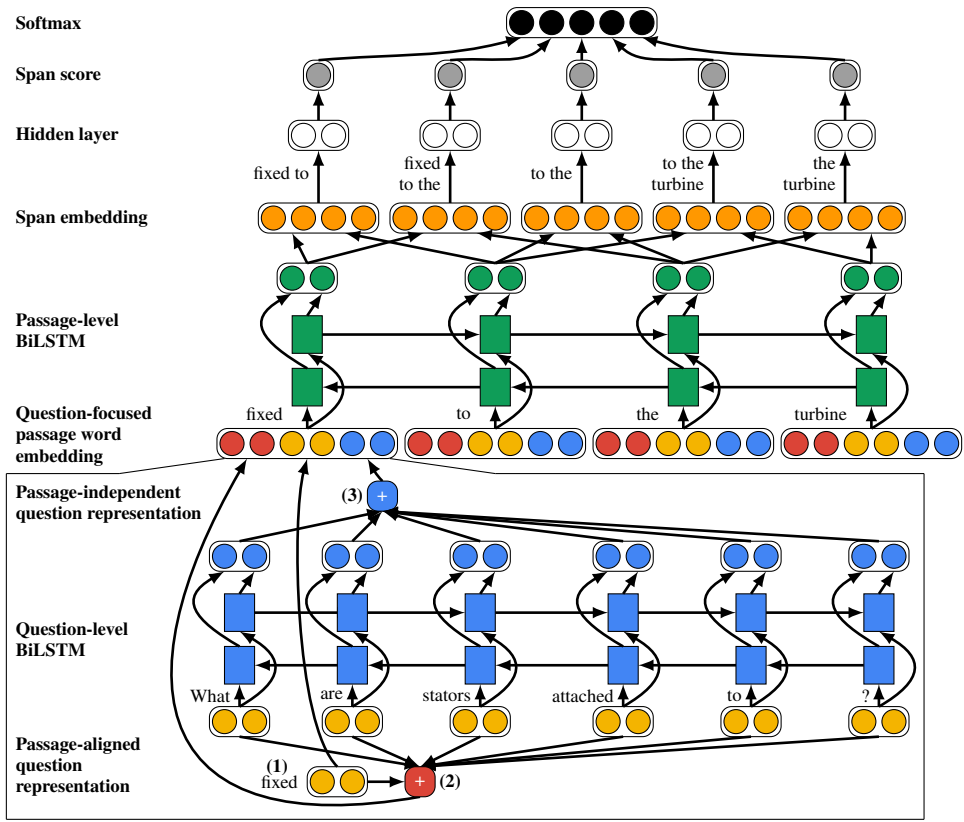
\begin{figure}[t]
\begin{centering}
\scalebox{1} {
\begin{tikzpicture}
\pword{0}{0}{fixed}
\lstmconnect{0}{0}{2.5}
\pword{2.5}{0}{to}
\lstmconnect{2.5}{0}{2.5}
\pword{5}{0}{the}
\lstmconnect{5}{0}{2.5}
\pword{7.5}{0}{turbine}
\pspan{0.1}{-0.7}{0}{2.5}{fixed to}{-0.8}{-165}{5}
\pspan{1.85}{-0.7}{0}{5}{fixed \\ to the}{-0.4}{-135}{5}
\pspan{3.6}{-0.7}{2.5}{5}{to the}{0}{-90}{90}
\pspan{5.35}{-0.7}{2.5}{7.5}{to the \\ turbine}{0.4}{-45}{175}
\pspan{7.1}{-0.7}{5}{7.5}{the \\ turbine}{0.8}{-15}{175}

\layer{3.15}{4.9}{5}{black} 

\node[anchor=west, align=left] at (-3.5, 5.1) {\textbf{Softmax}};
\node[anchor=west, align=left] at (-3.5, 4.4) {\textbf{Span score}};
\node[anchor=west, align=left] at (-3.5, 3.6) {\textbf{Hidden layer}};
\node[anchor=west, align=left] at (-3.5, 2.5) {\textbf{Span embedding}};
\node[anchor=west, align=left] at (-3.5, 1) {\textbf{Passage-level} \\ \textbf{\bilstm}};
\node[anchor=west, align=left] at (-3.5, -0.4) {\textbf{Question-focused} \\ \textbf{passage word } \\ \textbf{embedding}};
\node[anchor=west, align=left] at (-3.5, -1.3) {\textbf{Passage-independent} \\ \textbf{question representation}};
\node[anchor=west, align=left] at (-3.5, -3.1) {\textbf{Question-level} \\ \textbf{\bilstm}};
\node[anchor=west, align=left] at (-3.5, -4.8) {\textbf{Passage-aligned} \\ \textbf{question} \\ \textbf{representation}};
\node at (0.1, -4.8) {\textbf{(1)}};
\node at (2.4, -5) {\textbf{(2)}};
\node at (1.1, -1.2) {\textbf{(3)}};

\draw[-latex, line width=1pt, in=-60, out=135] (1.5, -1.2+0.2) to (1.3, -0.7);
\layercolorbox{0.4}{1.5-0.2}{-1.2-0.2}{1}{g-blue}
\node at (1.5, -1.2) {\textcolor{white}{+}};

\qword{-0.9}{-3.7}{What}
\lstmconnect{-0.9}{-3.7}{1.7}
\qword{0.6}{-3.7}{are}
\lstmconnect{0.6}{-3.7}{1.7}
\qword{2.3}{-3.7}{stators}
\lstmconnect{2.3}{-3.7}{1.9}
\qword{4.2}{-3.7}{attached}
\lstmconnect{4.2}{-3.7}{1.7}
\qword{5.9}{-3.7}{to}
\lstmconnect{5.9}{-3.7}{1.7}
\qword{7.6}{-3.7}{?}

\draw (-3.5,-5.5) -- (8.7, -5.5);
\draw (8.7,-5.5) -- (8.7, -0.9);
\draw (-3.5,-5.5) -- (-3.5, -0.9);
\draw (-3.5,-0.9) -- (-2, -0.9);
\draw (-2, -0.9) -- (-0.6, -0.7);
\draw (3,-0.9) -- (8.7, -0.9);
\draw (3,-0.9) -- (1.6, -0.7);

\draw[-latex, line width=1pt] (0.5+0.8, -5) to (2-0.2, -5); 
\node[anchor=east, align=right] at (0.5, -5) {fixed}; 
\layer{0.5}{-5-0.2}{2}{g-yellow} 
\draw[-latex, line width=1pt, in=-105, out=135, looseness=0.5] (0.5+0.4, -5+0.2) to (0.5, -0.7); 

\draw[-latex, line width=1pt, in=-120, out=-170, looseness=1.8] (2, -5-0.2) to (-0.3, -0.7);
\layercolorbox{0.4}{2-0.2}{-5-0.2}{1}{g-red}
\node at (2, -5) {\textcolor{white}{+}};

\end{tikzpicture}
}
\caption{
A visualization of \ourmodel, where the question is \textit{``What are the stators attached to?''} and the passage is \textit{``\dots fixed to the turbine \dots''}.
The model constructs question-focused passage word embeddings by concatenating \textbf{(1)} the original passage word embedding, \textbf{(2)} a passage-aligned representation of the question, and \textbf{(3)} a passage-independent representation of the question shared across all passage words.
We use a \bilstm~over these concatenated embeddings to efficiently recover embedding representations of all possible spans, which are then scored by the final layer of the model.}
\label{fig:span}
\end{centering}
\end{figure}

\section{Experimental Setup}
\label{sec:exp_setup}
We represent each of the words in the question and document using 300 dimensional GloVe embeddings trained on a corpus of $840{\rm bn}$ words \citep{pennington2014glove}.
These embeddings cover $200{\rm k}$ words and all out of vocabulary (OOV) words are projected onto one of $1{\rm m}$ randomly initialized $300{\rm d}$ embeddings.
We couple the input and forget gates in our LSTMs, as described in \cite{greff2015lstm}, and we use a single dropout mask to apply dropout across all LSTM time-steps as proposed by \cite{gal2015theoretically}. Hidden layers in the feed forward neural networks use rectified linear units \citep{relu}.
Answer candidates are limited to spans with at most 30 words.

To choose the final model configuration, we ran grid searches over: the dimensionality of the LSTM hidden states; the width and depth of the feed forward neural networks; dropout for the LSTMs; the number of stacked LSTM layers $(1,2,3)$; and the decay multiplier $[0.9, 0.95, 1.0]$ with which we multiply the learning rate every $10{\rm k}$ steps.
The best model uses $50 {\rm d}$ LSTM states; two-layer \bilstm s for the span encoder and the passage-independent question representation; dropout of $0.1$ throughout; and a learning rate decay of $5\%$ every $10{\rm k}$ steps.

All models are implemented using TensorFlow\footnote{\url{www.tensorflow.org}} and trained on the \squad~training set using the ADAM \citep{kingma2014adam} optimizer with a mini-batch size of $4$ and trained using $10$ asynchronous training threads on a single machine.

\section{Results}\label{sec:results}
We train on the $80{\rm k}$ (question, passage, answer span) triples in the \squad~training set and report results on the $10{\rm k}$ examples in the \squad~development and test sets.

All results are calculated using the official \squad~evaluation script, which reports exact answer match and F1 overlap of the unigrams between the predicted answer and the closest labeled answer from the $3$ reference answers given in the \squad~development set.

\subsection{Comparisons to other work}
Our model with recurrent span representations (\ourmodel) is compared to all previously published systems
\footnote{\label{ftnt:leaderboard}As of submission, other unpublished systems are shown on the \squad~leaderboard, including \textit{Match-LSTM with Ans-Ptr (Boundary+Ensemble)}, \textit{Co-attention}, \textit{r-net}, \textit{Match-LSTM with Bi-Ans-Ptr (Boundary)}, \textit{Co-attention old}, \textit{Dynamic Chunk Reader}, \textit{Dynamic Chunk Ranker with Convolution layer}, \textit{Attentive Chunker}.}.
\cite{rajpurkar:2016} published a logistic regression baseline as well as human performance on the \squad~task.
The logistic regression baseline uses the output of an existing syntactic parser both as a constraint on the set of allowed answer spans, and as a method of creating sparse features for an answer-centric scoring model.
Despite not having access to any external representation of linguistic structure, \ourmodel~achieves an error reduction of more than $50\%$ over this baseline, both in terms of exact match and F1, relative to the human performance upper bound.
\begin{table}[htp]
\begin{center}
\begin{tabularx}{0.6\textwidth}{ l c c c c c c}
 & \multicolumn{2}{c}{Dev}  & \multicolumn{2}{c}{Test} \\
 \midrule
 System & EM & F1 & EM & F1\\
\midrule
Logistic regression baseline & 39.8 & 51.0 & 40.4 & 51.0 \\
Match-LSTM (Sequence) & 54.5 & 67.7 & 54.8 & 68.0 \\
Match-LSTM (Boundary) & 60.5 & 70.7 & 59.4 & 70.0 \\
\ourmodel & 66.4 & 74.9 & 67.4 & 75.5 \\
Human & 81.4 & 91.0 & 82.3 & 91.2 \\
\end{tabularx}
\end{center}
\caption{Exact match (EM) and span F1 on \squad.}
\label{tab:results}
\end{table}

More closely related to \ourmodel~is the \textit{boundary model} with Match-LSTMs and Pointer Networks by \cite{wang2016machine}.
Their model similarly uses recurrent networks to learn embeddings of each passage word in the context of the question, and it can also capture interactions between endpoints, since the end index probability distribution is conditioned on the start index.
However, both training and evaluation are greedy, making their system susceptible to search errors when decoding.
In contrast, \ourmodel~can efficiently and explicitly model the quadratic number of possible answers, which leads to a $14\%$ error reduction over the best performing Match-LSTM model.

\subsection{Model Variations}
\label{sec:ablations}
We investigate two main questions in the following ablations and comparisons.
(1) How important are the two methods of representing the question described in Section~\ref{sec:pstar}?
(2) What is the impact of learning a loss function that accurately reflects the span prediction task?
\paragraph{Question representations}
Table~\ref{tab:ablations} shows the performance of \ourmodel~when either of the two question representations described in Section~\ref{sec:pstar} is removed.
The passage-aligned question representation is crucial, since lexically similar regions of the passage provide strong signal for relevant answer spans.
If the question is only integrated through the inclusion of a passage-independent representation, performance drops drastically.
The passage-independent question representation over the \bilstm~is less important, but it still accounts for over $3\%$ exact match and F1.
The input of both of these components is analyzed qualitatively in Section~\ref{sec:analysis}.
\begin{table}[htp]
\begin{center}
\subfloat[Ablation of question representations.]{
\begin{tabular}{l c c}
Question representation & EM & F1 \\
\midrule
Only passage-independent & 48.7 & 56.6   \\
Only passage-aligned &  63.1 & 71.3 \\
\ourmodel & 66.4 & 74.9 \\
\end{tabular}
\label{tab:ablations}
}
~~~~~~
\subfloat[Comparisons for different learning objectives given the same passage-level \bilstm.]{
\begin{tabular}{l c c}
Learning objective & EM & F1 \\
\midrule
Membership prediction & 57.9 & 69.7 \\
BIO sequence prediction &  63.9 & 73.0  \\
Endpoints prediction & 65.3 & 75.1 \\
Span prediction w/ log loss & 65.2 & 73.6 \\
\end{tabular}
\label{tab:labels}
}
\end{center}
\label{tab:variations}
\caption{Results for variations of the model architecture presented in Section~\ref{sec:model}.}
\end{table}

\paragraph{Learning objectives}
Given a fixed architecture that is capable of encoding the input question-passage pairs, there are many ways of setting up a learning objective to encourage the model to predict the correct span.
In Table~\ref{tab:labels}, we provide comparisons of some alternatives (learned end-to-end) given only the passage-level \bilstm~from \ourmodel. In order to provide clean comparisons, we restrict the alternatives to objectives that are trained and evaluated with exact decoding.

The simplest alternative is to consider this task as binary classification for every word (\textit{Membership prediction} in Table~\ref{tab:labels}). In this baseline, we optimize the logistic loss for binary labels indicating whether passage words belong to the correct answer span.
At prediction time, a valid span can be recovered in linear time by finding the maximum contiguous sum of scores.

\cite{li2016dataset} proposed a sequence-labeling scheme that is similar to the above baseline (\textit{BIO sequence prediction} in Table~\ref{tab:labels}).
We follow their proposed model and learn a conditional random field (CRF) layer after the passage-level \bilstm~to model transitions between the different labels.
At prediction time, a valid span can be recovered in linear time using Viterbi decoding, with hard transition constraints to enforce a single contiguous output.

We also consider a model that independently predicts the two endpoints of the answer span (\textit{Endpoints prediction} in Table~\ref{tab:labels}).
This model uses the softmax loss over passage words during learning.
When decoding, we only need to enforce the constraint that the start index is no greater than the end index.
Without the interactions between the endpoints, this can be computed in linear time.
Note that this model has the same expressivity as \ourmodel~if the span-level FFNN were removed.

Lastly, we compare with a model using the same architecture as \ourmodel~but is trained with a binary logistic loss rather than a softmax loss over spans (\textit{Span prediction w/ logistic loss} in Table~\ref{tab:labels}).

The trend in Table~\ref{tab:labels} shows that the model is better at leveraging the supervision as the learning objective more accurately reflects the fundamental task at hand: determining the best answer span.

First, we observe general improvements when using labels that closely align with the task.
For example, the labels for \textit{membership prediction} simply happens to provide single contiguous spans in the supervision.
The model must consider far more possible answers than it needs to (the power set of all words).
The same problem holds for \textit{BIO sequence prediction}-- the model must do additional work to learn the semantics of the BIO tags.
On the other hand, in \ourmodel, the semantics of an answer span is naturally encoded by the set of labels.

Second, we observe the importance of allowing interactions between the endpoints using the span-level FFNN. \ourmodel~outperforms the \textit{endpoint prediction} model by $1.1$ in exact match,
The interaction between endpoints enables \ourmodel~to enforce consistency across its two substructures.
While this does not provide improvements for predicting the correct \textit{region} of the answer (captured by the F1 metric, which drops by 0.2), it is more likely to predict a clean answer span that matches human judgment exactly (captured by the exact-match metric).


\newcolumntype{L}{>{\raggedright\arraybackslash}m{\textwidth}}
\newcommand{\hlwt}[2]{{\sethlcolor{#1}\hl{#2}}}

\definecolor{e1p1}{cmyk}{0.4, 0, 0, 0} 

\section{Analysis}
\label{sec:analysis}

\begin{figure}[hbp]
\begin{center}
\begin{minipage}{.3\textwidth}
\includegraphics[width=\textwidth]{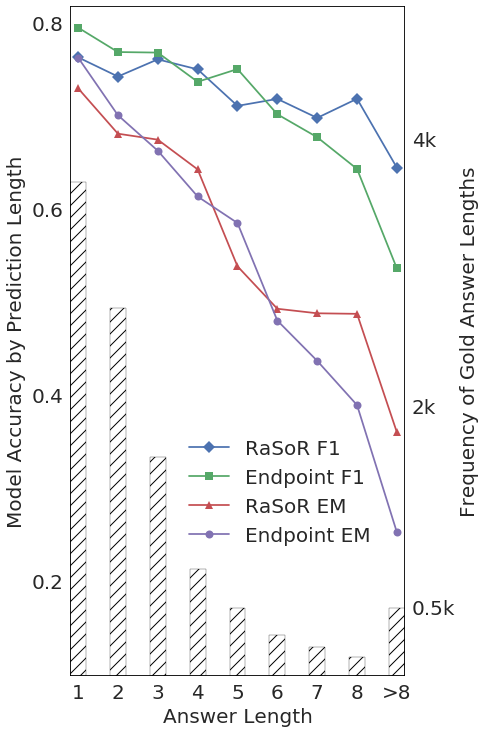}
\caption{F1 and Exact Match (EM) accuracy of \ourmodel~and the endpoint predictor baseline over different prediction lengths.}
\label{fig:len_analysis}
\end{minipage}~~~~~~~%
\begin{minipage}{.65\textwidth}
\includegraphics[width=\textwidth]{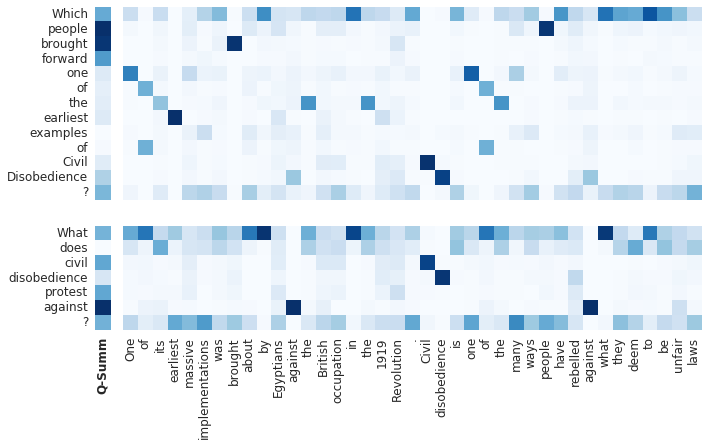}
\caption{Attention masks from \ourmodel. Top predictions for the first example are 'Egyptians', 'Egyptians against the British', 'British'.  Top predictions for the second are 'unjust laws', 'what they deem to be unjust laws', 'laws'.}
\label{fig:attention}
\end{minipage}

\end{center}
\end{figure}

Figure~\ref{fig:len_analysis} shows how the performances of \ourmodel~and the endpoint predictor introduced in Section~\ref{sec:ablations} degrade as the lengths of their predictions increase.
It is clear that explicitly modeling interactions between end markers is increasingly important as the span grows in length.

Figure~\ref{fig:attention} shows attention masks for both of \ourmodel's question representations.
The passage-independent question representation pays most attention to the words that could attach to the answer in the passage (``brought'', ``against'') or describe the answer category (``people'').
Meanwhile, the passage-aligned question representation pays attention to similar words.
The top predictions for both examples are all valid syntactic constituents, and they all have the correct semantic category. 
However, \ourmodel~assigns almost as much probability mass to it's incorrect third prediction ``British'' as it does to the top scoring correct prediction ``Egyptian''. 
This showcases a common failure case for \ourmodel, where it can find an answer of the correct type close to a phrase that overlaps with the question -- but it cannot accurately represent the semantic dependency on that phrase.

\section{Conclusion}
We have shown a novel approach for perform extractive question answering on the \squad~dataset by explicitly representing and scoring answer span candidates.
The core of our model relies on a recurrent network that enables shared computation for the shared substructure across span candidates.
We explore different methods of encoding the passage and question, showing the benefits of including both passage-independent and passage-aligned question representations.
While we show that this encoding method is beneficial for the task, this is orthogonal to the core contribution of efficiently computing span representation.
In future work, we plan to explore alternate architectures that provide input to the recurrent span representations.

\bibliography{main}
\bibliographystyle{iclr2017_conference}

\end{document}